\documentclass[lettersize,journal]{IEEEtran}
	
\IEEEoverridecommandlockouts

\usepackage{amsmath,amsfonts}
\usepackage{algorithmic}
\usepackage{algorithm}
\usepackage{array}
\usepackage[caption=false,font=normalsize,labelfont=sf,textfont=sf]{subfig}
\usepackage{textcomp}
\usepackage{stfloats}
\usepackage{multirow}
\usepackage{booktabs}
\usepackage{url}
\usepackage{verbatim}
\usepackage{graphicx}
\usepackage{cite}
\usepackage{graphicx}
\usepackage{etoolbox}
\usepackage{diagbox}
\usepackage{makecell}
\usepackage[colorlinks,
            linkcolor=blue,
            anchorcolor=blue,
            citecolor=blue]{hyperref}

\newenvironment{miao}{\begingroup}{\endgroup}
\newcommand{\miaot}[1]{#1}

\begin{document}

\title{Alignment Is All You Need For X-to-4D Generation}

\author{

Qiaowei Miao,~\IEEEmembership{Student Member,~IEEE,}
Kehan Li,
Yawei Luo$^*$,
Yi Yang,~\IEEEmembership{Fellow,~IEEE}
\thanks{This work was supported by National Natural Science Foundation of China (62293554, U2336212), "Pioneer" and "Leading Goose" R\&D Program of Zhejiang (2025C02022, 2024C01161), Zhejiang Provincial Natural Science Foundation of China (LZ24F020002), Ningbo Innovation "Yongjiang 2035" Key Research and Development Programme (2024Z292), and Young Elite Scientists Sponsorship Program by CAST (2023QNRC001). The author gratefully acknowledges the support of Zhejiang University Education Foundation Qizhen Scholar Foundation. This research was supported by HUAWEI's Al Hundred Schools Program and was carried out using the Ascend AI technology stack.}

\thanks{Qiaowei Miao, Kehan Li, Yawei Luo, and Yi Yang are with the Zhejiang University, Hangzhou 310027, China (e-mail: qiaoweimiao@zju.edu.cn; kehan.li@zju.edu.cn;yaweiluo@zju.edu.cn;
yangyics@zju.edu.cn).}
\thanks{ Yawei Luo is the corresponding author.}
}

\markboth{Journal of \LaTeX\ Class Files,~Vol.~14, No.~8, August~2021}%
{Shell \MakeLowercase{\textit{et al.}}: A Sample Article Using IEEEtran.cls for IEEE Journals}


\maketitle

\begin{abstract}
Generative diffusion models excel at synthesizing high-quality images, videos, and 3D content under multimodal control. However, arbitrary user-defined modality-to-4D (X-to-4D) generation remains challenging due to the high cost of constructing diverse datasets and the limited scalability of existing methods. This paper presents Align4D, a flexible framework that translates any-modal input into coherent video–3D pairs, using video to guide 4D motion and 3D data to shape 4D geometry. Align4D introduces three key techniques: (1) \textbf{Object Distance Alignment}, which searches Video-Aligned and Multiview-Aligned Object Distances (VAOD/MAOD) respectively, to reconcile 4D renderings to video and the priors of multiview diffusion models; (2) \textbf{Motion-Geometry Joint Alignment}, which constrains known and unknown views through synchronized video and 3D inputs, ensuring consistent 4D generation; and (3) \textbf{Asynchronous Optimization}, which decouples Gaussian attribute and deformation network training to enhance motion and geometry fidelity. We further propose the X4D dataset, integrating prompt, image, video, and 3D data for benchmarking. Experiments on X4D and Consistent4D demonstrate that Align4D achieves state-of-the-art quality and consistency in X-to-4D generation. Project page: \url{https://miaoqiaowei.github.io/Align4D/}.
\end{abstract}


\section{Introduction}

\IEEEPARstart{4}{D} content generation is essential for creating realistic and dynamic content in applications such as virtual reality, the metaverse, computer games, and cinematic visual effects. However, existing 4D generation methods~\cite{AYG, 4d-fy, 4dgen, make-a-video-3d, Consistent4d, Animate124, dreamgaussin4d, ren2024l4gm, pan2024efficient4d, wu_sc4d_2024} typically focus on single-modal inputs, which limits their generative flexibility and capabilities. Text-to-4D methods~\cite{AYG, 4d-fy, 4dgen} provide broad accessibility, but the generated objects often lack a strong sense of realism and purpose. Image-to-4D methods~\cite{yang_diffusionmbox2_2024, dreamgaussin4d, pang2024disco4d} enhance appearance quality but fail to effectively provide motion guidance. Video-to-4D methods~\cite{wang2024vidu4d, ren2024l4gm,pan2024efficient4d, wu_sc4d_2024, Consistent4d} improve motion realism but compromise structural consistency. Meanwhile, 3D-to-4D methods~\cite{Animate124} preserve spatial geometry but lack temporal coherence. Recent efforts~\cite{xie_sv4d_2024, liang_diffusion4d_2024} explore constructing datasets to train diffusion models for generating multiview and multi-timestamp images. However, scaling these datasets to enable multimodal inputs remains infeasible due to the prohibitive data collection and computational resource requirements. These limitations highlight the urgent need for a flexible and unified framework that enables arbitrary modality-to-4D (X-to-4D) generation.

\begin{figure}[t]
    \centering
    \includegraphics[width=0.88\linewidth]{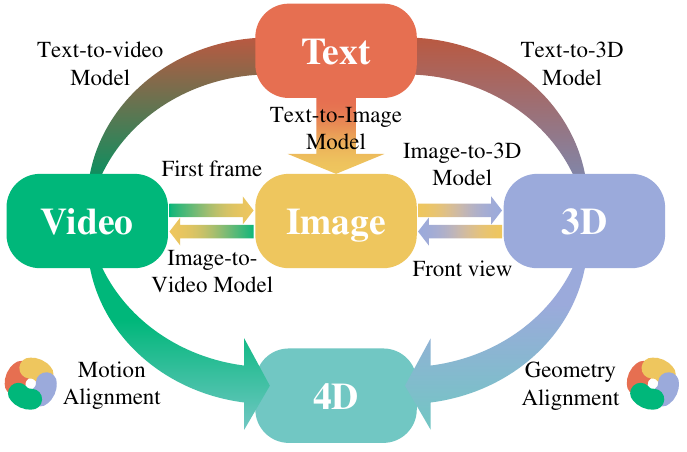}
    \caption{ \textbf{Align4D transforms an arbitrary input into a coherent \textit{Video-3D pair} by sequentially leveraging multiple off-the-shelf models.} Within this X-to-4D generation framework, Align4D focuses on rigorously aligning the  4D object’s temporal motion with the video prior and its spatial geometry with the 3D representation, thereby achieving powerful synthesis capabilities.}
    \label{fig:idea}
\end{figure}

\begin{figure*}[tp] 
    \centering
  \includegraphics[width=0.90\textwidth]{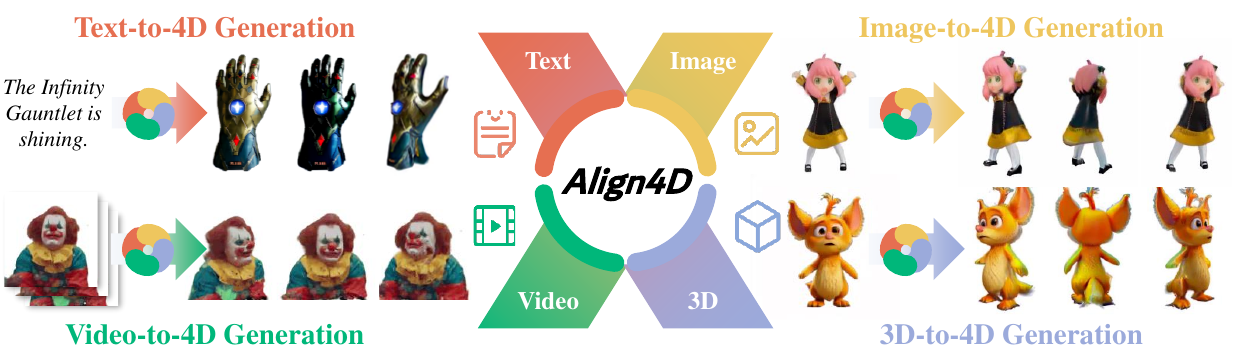}
  \caption{
  \begin{miao}
      \textbf{Align4D is a novel X-to-4D framework that enables users to input text, images, videos, or 3D objects to generate 4D targets.} Align4D transforms arbitrary modality inputs into corresponding video-3D pairs, synchronizing the motion of the generated 4D objects with the videos and aligning their structures with the 3D data through matched object distances , achieving dynamic objects with high temporal dynamics and precise geometry.
  \end{miao}
  }
  \label{fig:teaser}
\end{figure*}

\begin{miao}

In this paper, we revisit the 4D generation task and present a key insight: \textbf{4D content synthesis can be intrinsically decoupled into 3D geometry generation and temporal motion generation.} Building on this idea, we propose \textbf{Align4D}, a unified framework that reformulates arbitrary-modality-guided 4D generation into a task of aligning 4D assets with synchronized video and 3D data through matched object distances. Instead of training an end-to-end model from scratch, our framework leverages the robust inference capabilities of off-the-shelf pretrained diffusion models to bridge the gap between input modalities and 4D outputs. As illustrated in Figure~\ref{fig:idea}, Align4D establishes flexible generative pathways where diversity input (text, image, video, or 3D) is converted into a coherent \textit{Video-3D pair} via video generation models~\cite{chen2023videocrafter1,videocrafter2,sora,Stablevideo} and 3D generation models~\cite{tang2024lgm,Crm,meshy_ai_2025}. 
Align4D then focuses on rigorously aligning the temporal motion of the 4D target with the video input, and ensuring its spatial geometry is consistent with the 3D representation. This dual alignment process is critical, as it guarantees the synthesized 4D results exhibit both action coherence and structural fidelity. This modular design effectively circumvents the data scarcity issue, thereby offering a generalizable and flexible solution for X-to-4D generation with powerful synthesis capabilities, as shown in Figure~\ref{fig:teaser}.

\end{miao}

However, unifying these disparate priors presents a significant challenge: \textbf{aligning the 4D output with both the motion posture from the generated video and the geometric structure from the generated 3D content.} Since the video and 3D priors are derived from independent models, they often exhibit spatial or temporal discrepancies. We identify two critical aspects of this alignment problem. First, for \textbf{Known Spatiotemporal Viewpoints}, the 4D target must align with the 3D geometry at the initial state and the video content across time. A major hurdle here is determining the appropriate object distance; existing methods~\cite{li2024dreammesh4d,wu_sc4d_2024,Consistent4d} often rely on manual, empirical settings, which lead to floating artifacts or distortions when applied to arbitrary inputs. Second, for \textbf{Unknown Spatiotemporal Viewpoints}, the model must maintain geometric consistency across unseen views while adhering to the motion dynamics observed in the video. Optimizing 4D targets using multiple diffusion priors simultaneously is notoriously unstable and time consuming~\cite{ren2024l4gm,dreamgaussin4d,Dreamfusion}, often resulting in degraded quality due to parameter sensitivity.

To tackle these challenges, Align4D introduces a novel alignment-driven optimization strategy designed to synthesize 4D objects with smooth motion and consistent geometry. First, we propose \textbf{Object Distance Alignment}, which automatically searches for the Video-Aligned Object Distance (VAOD) to match front-view renderings with video frames, and the Multiview-Aligned Object Distance (MAOD) to calibrate the 4D model with multiview diffusion priors. Furthermore, we develop a \textbf{Motion-Geometry Joint Alignment (MGJA)} module. For known viewpoints, we utilize VAOD to strictly align the 4D model with the input video. For unknown viewpoints, rather than relying on unstable multi-model optimization, we employ a single multiview diffusion model guided by MAOD and condition it on the video frames to transfer motion and geometry priors to non-front views. To ensure robust convergence, we introduce an \textbf{Asynchronous Optimization} strategy that refines geometry and motion parameters separately. Extensive experiments on our newly collected X4D dataset and the Consistent4D dataset~\cite{Consistent4d} demonstrate that Align4D effectively harmonizes conflicting priors, generating high-fidelity 4D targets that often surpass the quality of the initial video-3D pairs.
Our contributions are as follows:
\begin{itemize}
    \item We propose an X-to-4D generation framework that supports arbitrary input modalities, including text, images, videos, and 3D data, as conditioning inputs to generate 4D assets.
    \item We introduce a novel object distance alignment method designed to search the matched object distances for aligning 4D renderings with video and the prior of a multiview diffusion model. Building on this foundation, we employ a single multiview diffusion model to jointly align the motion and geometry of 4D renderings with video and 3D data in an asynchronous way. 
    \item We construct X4D, a first quadruple dataset to benchmark X-to-4D generation capabilities via inputs generated by pretrained diffusion models.
    \item Extensive experiments demonstrate that Align4D excels in generating 4D objects, producing fine textures, precise geometry, and seamless motion.
\end{itemize}

\section{Related works}

\textbf{Generative diffusion models.} They revolutionize the landscape of visual generation, demonstrating remarkable performance across tasks such as image, video, and 3D content generation~\cite{miao2025advances4dgenerationsurvey}. 
(a) Image generation models~\cite{rombach2022high,saharia2022photorealistic,i2i1, i2i2} leverage their impressive text-guided generative capabilities to produce high-resolution and highly creative images, sparking significant interest in extending diffusion models to video and 3D generation domains. 
(b) Video generation models~\cite{ Alignyourlatents,ge2023preserve, Emuvideo, Animatediff, Imagenvideo, Text2video-zero, Stablevideo,CVG,zhang2025video2roleplay,vmtion}, including text-to-video and image-to-video generation, garner increasing attention. For instance, text-to-video models~\cite{make-a-video,Alignyourlatents} rely on large-scale, high-quality text-to-video datasets for training, enabling a deeper understanding of verbs and generating rich, creative sequences of temporally coherent video frames. On the other hand, image-to-video models~\cite{Animatediff, Stablevideo} infer the subsequent actions of a target object based solely on a given initial frame. However, these models lack flexible control over the generated actions. 
(c) 3D generation models~\cite{mvdream,Zero123++,wonder3d,Crm,t23d,T3D, li2024dreamtexture, min2024entangled,min2024epipolar} allow users to set text or images as control conditions to generate static 3D targets based on meshes or 3D Gaussian representations. These models achieve robust generalization through large-scale training datasets and can rapidly generate 3D objects for unseen inputs using inference alone. The rapid advancements in image, video, and 3D diffusion models lay a solid foundation for constructing an X-to-4D generation framework.

\begin{miao}

\textbf{4D Generation.} The goal of 4D generation is to synthesize dynamic 3D objects with consistent geometry and motion across views. However, existing approaches still face several critical limitations across different dimensions.
Regarding \textbf{control modalities}, current methods primarily rely on single-modal inputs: while text-driven methods~\cite{4d-fy,make-a-video-3d} and image-to-4D pipelines~\cite{dreamgaussin4d} offer partial multimodal support, dedicated video-to-4D~\cite{ren2024l4gm,wu_sc4d_2024,Stag4d} and 3D-to-4D~\cite{Animate124} methods cannot jointly accommodate text, image, 3D, and video as unified inputs.
Concerning \textbf{representations}, NeRF-based approaches~\cite{Dream-in-4D,4d-fy,make-a-video-3d,4dgen} provide smooth temporal modeling but suffer from prohibitive training times. Conversely, Gaussian Splatting variants~\cite{dreamgaussin4d,AYG,Stag4d,quan2026particlegs,miao2026frequency,xu2026langfield4d} improve efficiency yet struggle to provide fine-grained motion control over thousands of Gaussians, compromising temporal coherence.
For \textbf{guidance and optimization}, many approaches rely on proprietary video diffusion models~\cite{make-a-video,Alignyourlatents} for motion guidance, which severely limits reproducibility. Open-source attempts~\cite{4d-fy,dreamgaussin4d} often suffer from inaccurate object distances and an entangled motion–geometry optimization process, leading to degraded 4D consistency. Furthermore, the reliance on optimization from random initialization via SDS contributes to slow convergence and prolonged generation times.
In contrast, our \textbf{Align4D} introduces a 3DGS-based 4D framework with a deformable motion field and novel multimodal alignment strategies. This design enables efficient generation while preserving both geometric fidelity and video-consistent motion.

\end{miao}
\vspace{-5pt}
\section{Methodology} 

\begin{figure*}[tph]
    \centering
    \includegraphics[width=0.85\linewidth]{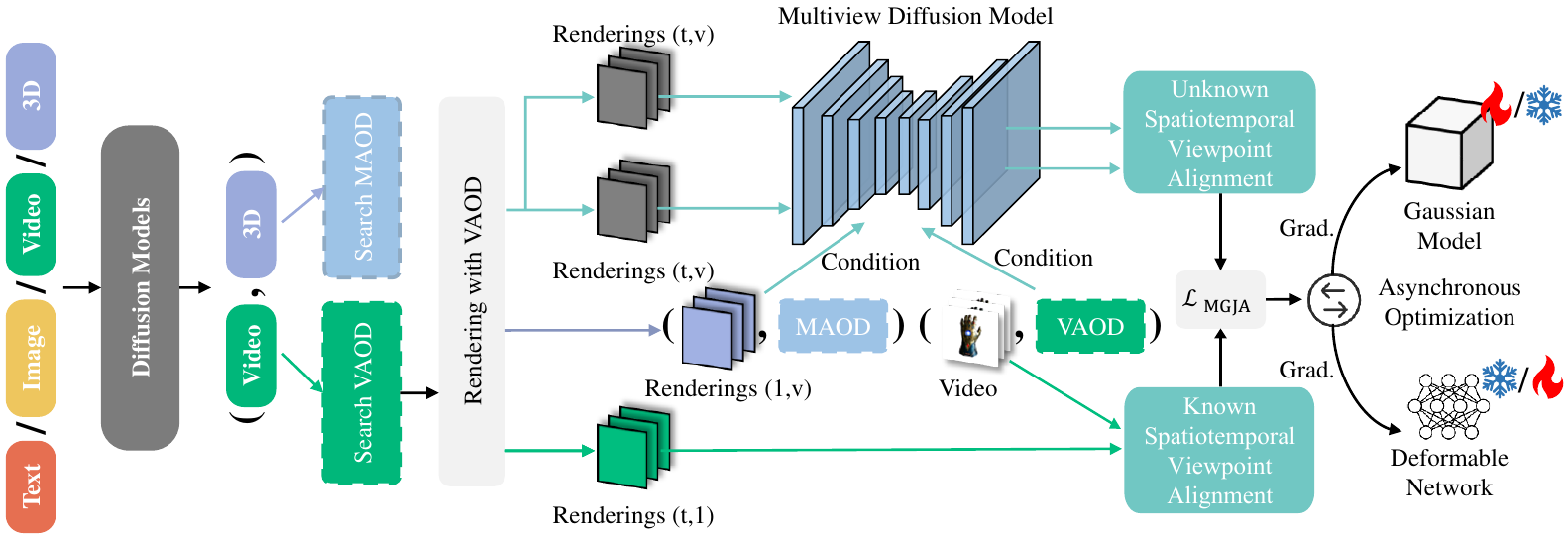}
    \caption{\textbf{Overview of the Align4D framework.} 
Given arbitrary input modalities, pretrained video and 3D diffusion models are used to construct a unified video–3D pair. We then search for two key object distances: the Video-Aligned Object Distance (VAOD), which aligns the 4D front-view renderings with the video, and the Multiview-Aligned Object Distance (MAOD), which aligns multiview geometry with the multiview diffusion prior~\cite{Zero-1-to-3}. 
Using these distances, Align4D performs Motion–Geometry Joint Alignment under both known and unknown spatiotemporal viewpoints. Finally, an asynchronous optimization stage refines the Gaussian representation and the deformation network to produce high-quality 4D content.}
    \label{fig:overall}
\end{figure*}

In this paper, we introduce Align4D, an X-to-4D generation framework that aligns motion with video priors and geometry with 3D inputs. Leveraging pretrained models, Align4D converts arbitrary modalities into unified video–3D pairs for 4D synthesis.
We first outline the preliminaries (Section~\ref{sec:pre}). To ensure accurate motion–geometry alignment, we propose object distance alignment, which identifies the video-aligned object distance and multiview-aligned object distance (Section~\ref{sec:FLA}). Based on these distances, we design a motion–geometry joint alignment module for both known and unknown spatiotemporal viewpoints (Section~\ref{sec:MGJA}). Finally, an asynchronous optimization strategy further refines the 4D representation (Section~\ref{sec:AO}), and we introduce the constructed multimodal dataset X4D (Section~\ref{sec:data}). Our framework is illustrated in Figure~\ref{fig:overall}.

\subsection{Preliminaries}\label{sec:pre}


\textbf{Score Distillation Sampling (SDS).}
SDS is a widely adopted mechanism in 3D and 4D generation~\cite{mvdream, Dreamfusion, dreamgaussin4d, Prolificdreamer}, enabling the transfer of 2D diffusion priors into 3D representations. 
During diffusion model training, a noised sample $x_\tau$ is produced by adding Gaussian noise $\epsilon\!\sim\!\mathcal{N}(0,\mathbf{I})$ to data $x$ at a randomly sampled timestep $\tau$, and the UNet $\phi$ is trained to predict the added noise $\epsilon$. 
Given a 3D scene representation with parameters $\theta$, its rendering $x=g(\theta)$ is treated as the diffusion input. The SDS gradient~\cite{Dreamfusion} is then computed as:
\begin{equation}
    \label{eq:sds_grad}
    \nabla_{\theta}\mathcal{L}_{\mathrm{SDS}}(x=g(\theta))
    = \mathbb{E}_{\tau,\epsilon}
    \left[w(\tau)\big(\hat{\epsilon}_{\phi}(\mathbf{z}, v, \tau)-\epsilon\big)
    \frac{\partial x}{\partial \theta}\right],
\end{equation}
where $\mathbf{z}$ is the latent corresponding to $x$, $v$ denotes conditioning signals, and $w(\tau)$ is the timestep-dependent weighting. 
This gradient is backpropagated through the differentiable renderer $g$ to update the 3D parameters~$\theta$.

\subsection{Object Distance Alignment}\label{sec:FLA}

\textbf{Theoretical analysis.}
Whether using explicit video supervision or SDS-based implicit supervision, consistent object-distance alignment is fundamental to 4D generation. In video-conditioned settings, the camera captures the target at a fixed object distance, whereas pretrained diffusion models are trained on images rendered with a canonical focal length but varying distances. This mismatch makes 4D generation challenging: the reconstructed 4D object must simultaneously match the video’s object distance and conform to the spatial priors encoded in the diffusion model. Thus, estimating both the video-aligned object distance and the multiview diffusion–aligned object distance becomes essential.

We introduce the pinhole camera model to illustrate how to calculate the object distance. For a given camera with a fixed focal length $f$, an image $x$ of a 4D target $\theta$ is captured at an object distance $d$. If the true physical width of the 4D target $\theta$ is $W$, and its projected width in image $x$ is $w$, then by similar triangles, we get:
\begin{equation}\label{eq:1}
 \frac{w}{f} = \frac{W}{d}, \quad d=\frac{W}{w}f.
\end{equation}
However, a 4D target isn't a physical entity, so its physical dimensions cannot be directly measured to derive the object distance. Therefore, we propose a search method to find the optimal object distance.
Based on Equation~\ref{eq:1}, the ratio of any object distance $d_s$ to the video-aligned object distance $d_v$ is equal to the ratio of the target size in the rendered image $w_s$ to the target size in the video frame $w_v$:
\begin{equation}\label{eq:3}
 \frac{d_v}{d_s}= \frac{w_s}{w_v}.
\end{equation}
Thus, by varying $d_s$ to generate rendered images $x_s$, we compute the corresponding widths $w_s$ and compare them with the target width $w_v$ in the video frame. When $w_s$ closely matches $w_v$, Equation~\ref{eq:3} indicates that $d_v = d_s$, \emph{i.e.}, $d_s$ is identified as the video-aligned object distance.

\begin{figure*}[tp]
    \centering
    \includegraphics[width=0.85\linewidth]{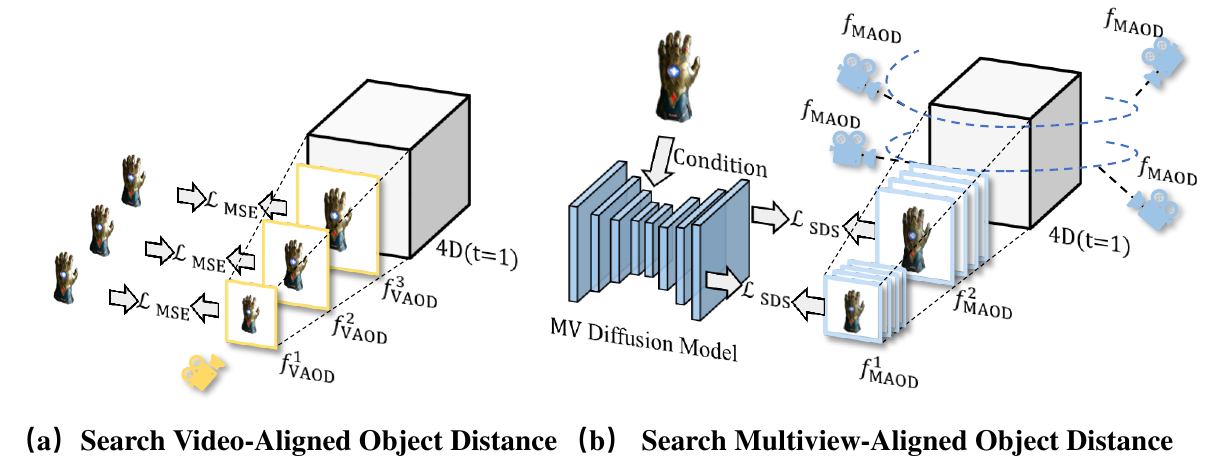}
    \caption{\textbf{Object distance alignment.}
    (a) We search for the Video-Aligned Object Distance (VAOD) to align the known front-view renderings of the 4D object with the input video. Specifically, we render front-view images under different object distances and compute $\mathcal{L}_{\text{MSE}}$ with respect to the video frames; the distance that yields the global minimum $\mathcal{L}_{\text{MSE}}$ is selected as the VAOD.
    (b) We then search for the Multiview-Aligned Object Distance (MAOD), which aligns multiview renderings with the geometry prior of the multiview diffusion model. Using the first frame as the condition, we render four orthogonal views and compute $\mathcal{L}_{\text{SDS}}$. The object distance corresponding to a local minimum of $\mathcal{L}_{\text{SDS}}$, and that is smaller than the VAOD, is selected as the MAOD.
    }
    \label{fig:search}
    \vspace{-10pt}
\end{figure*}


For the object distance associated with the multiview diffusion model, the matched value should yield rendered inputs whose output distribution matches the model’s training-time distribution. Since diffusion models enforce alignment between their outputs and a Gaussian prior during training, we can assess how well a candidate object distance aligns with the training distribution by fixing condition and varying only the object distance. In this setting, the similarity between the UNet-predicted noise and the Gaussian prior serves as an indicator of distance alignment. Motivated by this observation, we treat the diffusion UNet as a distributional discriminator and use the SDS loss—an effective proxy for the KL divergence between the model’s output distribution and the prior~\cite{Dreamfusion}—to evaluate the degree of alignment at each object distance. This allows us to reliably search for the multiview diffusion model–aligned object distance. Building on this theoretical basis, we then develop concrete search procedures for the Video-Aligned Object Distance (VAOD) and the Multiview-Aligned Object Distance (MAOD).

\textbf{Video-Aligned Object Distance (VAOD).}
To determine the object distance that best aligns the 4D model with the input video, we search for the Video-Aligned Object Distance (VAOD). As shown in Figure~\ref{fig:search} (a), we initialize the 4D representation $\theta$ using the provided 3D model $\psi$ at $t=1$. We then sweep the object distance $d' \in \mathcal{D}=[d_{\min}, d_{\max}]$ and render the front-view images $\{x_{\theta_1}^{d'}\}$. Each rendering is compared with the first video frame $I_1$ using the Mean Squared Error (MSE), and the distance yielding the global minimum is selected as:
\begin{equation}
    d_{\mathrm{VA}}
    = \arg\min_{d'} \|x_{\theta_1}^{d'} - I_1\|_2^2 .
    \label{eq:focal}
\end{equation}

\begin{figure}[tp]
    \centering
    \includegraphics[width=1\linewidth]{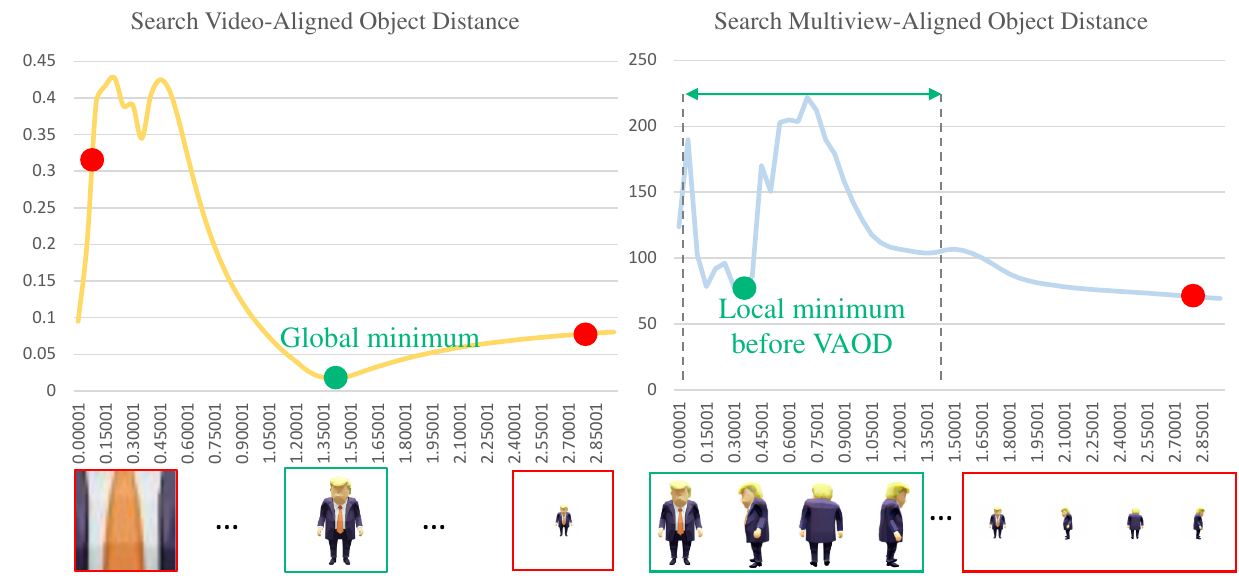}
    \caption{\textbf{Searching strategies for video-aligned object distance and multiview-aligned object distance.}}
    \vspace{-10pt}
    \label{fig:whysearch}
\end{figure}

The rationale for choosing the global minimum is illustrated in Figure~\ref{fig:whysearch}. When $d'$ is too small, the object becomes overly magnified, producing saturated renderings with uniformly low MSE. As $d'$ increases, the object gradually becomes fully visible, and the MSE decreases until reaching the best geometric and motion correspondence. Beyond this point, further increases in $d'$ cause the object to shrink and eventually vanish, driving the rendered image toward a white background and increasing the MSE toward its upper bound.

\textbf{Multiview-Aligned Object Distance (MAOD).}  Selecting a matched object distance is crucial when using the SDS method with multiview diffusion models to provide geometric priors. This necessity stems from the limited scope of pretrained data in these models, often resulting in the classification of images for 4D generation as out-of-distribution data, thereby making generative outcomes heavily dependent on the diffusion model's ability to generalize. The aligned object distance, as a key control parameter, significantly enhances the model's capacity to adapt to out-of-distribution data and strengthens its ability to provide geometric priors within the SDS framework, ultimately improving the quality of 4D generation.

To find the optimal object distance parameter, we search for the Multiview-Aligned Object Distance (MAOD) to align multiview renderings of 4D objects with the multiview diffusion model's prior. As illustrated in Figure~\ref{fig:search} (b), we obtain orthogonal four-view rendered images corresponding to azimuth angles \([-90^\circ, 0^\circ, 90^\circ, 180^\circ]\). Within the range \([d_{\text{min}}, d_{\text{max}}]\), we render a set of images \(\left\{ \left\{ x_\mathrm{\theta_{1}}^{c,d'} \right\}_{c \in C} \right\}_{d' \in \mathcal{D}}\), where \(C\) denotes the coordinates. We then use the video frame \(I_{1}\) at time step \(t=1\) as the image control condition. It is important to note that \(I_{1}\) is equivalent to the 4D front rendered image \(x_\mathrm{\theta_{1}}^{d_{\text{VA}}}\) corresponding to the VAOD. Next, the four orthogonal rendered images at each object distance \(d'\) are used as inputs to compute the modified SDS loss \(\mathcal{L}_{\text{SDS}}\) for searching:
\begin{equation} \label{eq:core}
\mathcal{L}_{\text{SDS}} = \frac{1}{|C| |\text{T}|} \sum_{c \in C} \sum_{\tau \in \text{T}} w(\tau)\left\| \epsilon_{\phi}(z_{\theta_{1}}^{c} ; I_{1}, c, d',\tau)-\epsilon \right\|^{2}_{2}, 
\end{equation}
where $z_{\theta_{1}}^{c}=\alpha_{\tau}x_{\theta_{1}}^{c}+\sigma_{\tau} \epsilon_{\text{fix}}$. We sample a noise latent \(\epsilon_{\text{fix}}\) and add it to each object distance and each viewpoint rendering \(x_{\theta_{1}}^{c}\) to calculate \(\mathcal{L}_{\text{SDS}}\). Here, \(c\) represents the coordinate of the rendering \(x_{\theta_{1}}^{c}\), \(\tau\) is the timestep of the multiview diffusion model. We use diffusion timesteps $\text{T}=\{700, 800, 900\}$.

\begin{figure}[tp]
    \centering
    \includegraphics[width=1\linewidth]{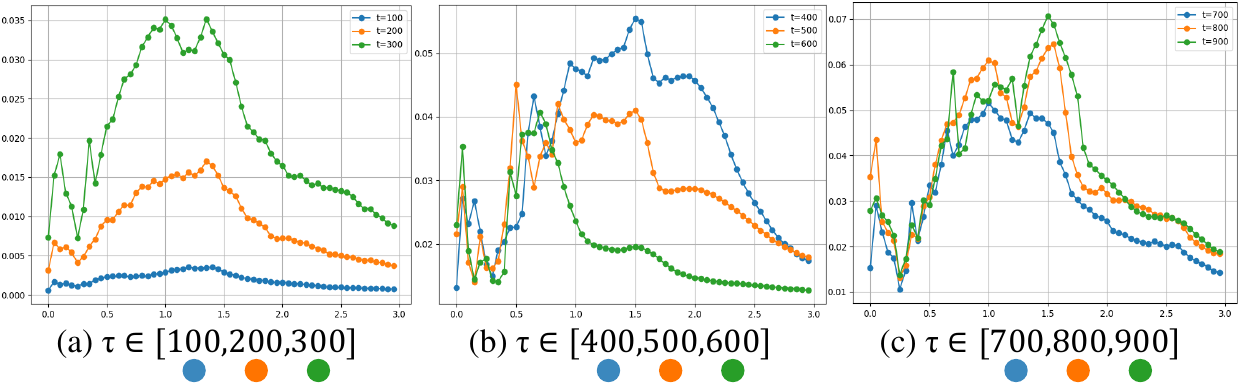}
        \caption{
        \miaot{
        \textbf{SDS loss versus object distance for different diffusion timesteps $\tau$.}
        (a) $\tau \in \{100,200,300\}$, the curves exhibit irregular and unstable variations. 
        (b) $\tau \in \{400,500,600\}$, a coarse decreasing–then–increasing trend emerges, but the minima remain inconsistent across distances. 
        (c) $\tau \in \{700,800,900\}$, the curves show strong consistency with a clear local minimum near the matched object distance.
        Averaging the SDS losses across multiple large-$\tau$ timesteps yields a stable and reliable estimate of the MAOD.
        }}
    \label{fig:tau}
\vspace{-10pt}
\end{figure}

\begin{miao}

The choice of the diffusion timestep $\tau$ is closely related to the noise schedule coefficients $w(\tau)$, which control the gradual noise injection in diffusion models—ensuring that early steps remain close to the data distribution while later steps approach a Gaussian prior~\cite{kingma2021variational,DBLP:journals/corr/abs-2304-13224}.
A smaller $\tau$ increases the mismatch between the predicted noise and the injected Gaussian noise, making the SDS loss less reliable for reflecting how well the diffusion prior aligns with the rendered inputs.
We provide the SDS–object-distance curves in Figure~\ref{fig:tau}. (a) When $\tau \in \{100,200,300\}$ display no stable pattern, resulting in unreliable MAOD estimation. (b) Intermediate timesteps $\tau \in \{400,500,600\}$ begin to exhibit a decreasing-then-increasing trend but remain inconsistent across distances. (c) In contrast, $\tau \in \{700,800,900\}$ produce highly consistent curves with a pronounced local minimum near the true object distance, enabling robust MAOD identification.
Therefore, we adopt $\tau \in \{700,800,900\}$ and use the averaged SDS loss across these timesteps, which reduces stochastic variability and yields stable, reliable MAOD estimates.
\end{miao}

Compared with the MSE loss used for VAOD, the SDS loss exhibits fundamentally different behavior. A well-trained diffusion model produces a lower SDS loss when the input aligns well with its learned prior. Consequently, within the distance range smaller than VAOD, using the video frame $I_{1}$ as the conditioning view yields a clear local minimum, which corresponds to the object distance most compatible with the multiview diffusion prior and is thus selected as the MAOD.
For distances larger than VAOD, the rendered object progressively disappears, causing the SDS loss to drop toward a trivial lower bound associated with a blank background, as illustrated in Figure~\ref{fig:search}. Although this global minimum is numerically smaller, it does not reflect meaningful geometric alignment. Our experiments show that the correct MAOD is given by the local minimum located to the left of VAOD, whereas the global minimum is an artifact caused by background-dominated renderings. More analysis is provided in Section~\ref{sec:why}.

\subsection{Motion-Geometry Joint Alignment}\label{sec:MGJA}

\textbf{Known Spatiotemporal Viewpoint Alignment.} 
First, we align the given video with the multi-time-step front views of the 4D target. For a generated video consisting of $\mathcal{T}$ frames, \(\left \{ I_{t} \right \}_{\mathcal{T}}\), we render the corresponding $\mathcal{T}$ frames under the front view $c'$ at the object distance \(d_{\text{VA}}\), \(\left \{ x_\mathrm{\theta_{t}}^{c'} \right \}_{\mathcal{T}}\). We calculate the MSE loss between them to inject motion into the 4D model:
\begin{equation}
    \mathcal{L}_{\mathrm{KSVA}} = \frac{1}{\mathcal{T}} \sum_{t=1}^{\mathcal{T}} || x_{\theta_{t}}^{c'} - I_{t} ||_{2}^{2} + || m_{\theta_{t}}^{c'} - M_{t} ||_{2}^{2} ,
    \label{eq:KSVA}
\end{equation}
where $\theta_{t}$ denotes  the 4D  model at time $t$, $m_{\theta_{t}}^{c'}$ represents the  alpha channel of $x_{\theta_{t}}^{c'}$ and $M_{t}$ represents the mask of frame $I_{t}$.


\textbf{Unknown Spatiotemporal Viewpoint Alignment.}
Beyond enforcing constraints on frontal actions, extending motion information to other unknown viewpoints remains a significant challenge. Previous approaches~\cite{dreamgaussin4d, 4d-fy, 4dgen, make-a-video-3d, AYG} attempt to impose constraints using SDS optimization via image diffusion models, multiview diffusion models, and video diffusion models. However, this indiscriminate stacking of SDS not only incurs substantial computational overhead but also fails to guarantee cohesive results. To address this issue, we revisit the optimization strategy of multiview diffusion models for SDS and propose the Motion-Geometry Joint Alignment (MGJA).

MGJA leverages a single multiview diffusion model, referencing both video and 3D data, to effectively transfer motion and geometric performance from frontal views to unknown targets in nonfrontal views and non-initial timesteps for 4D objects. We utilize one multiview diffusion model~\cite{Zero-1-to-3} to simultaneously align 4D renderings with video motion and 3D geometry.
The motion and posture of a 4D object in unknown views need to align with the corresponding video frame at each time step. Specifically, as shown in Figure~\ref{fig:overall}, for each time step \(t\), we randomly select \(N\) viewpoints. These \(N\) viewpoints are used to render a set of corresponding images from the 4D model at the video-aligned object distance \(d_{\text{VA}}\). Subsequently, we utilize \(\mathcal{L}_{\mathrm{mot}}\) to transfer the motion information from the front view to these rendered viewpoints:
\begin{equation}
\mathcal{L}_{\mathrm{mot}} = \frac{1}{N}\sum_{n=1}^{N}w(\tau)\left\|\epsilon_{\phi}(\alpha_{\tau}x_{\theta_{t}}^{c_{n}}+\sigma_{\tau} \epsilon; I_{t} , c_{n},  d_{\text{VA}}, \tau)-\epsilon\right\|^{2}_{2},
\label{eq:mot}
\end{equation}
where $\epsilon$ is the randomly sampled noise, \(\phi\) represents the U-Net of the multiview diffusion model, and \(w(\tau)\) denotes the weight coefficient related to the timestep. The variable \(\tau\) is a randomly sampled timestep within the multiview diffusion model. $I_{t}$ is the  $t$-th video frame, and $c_{n}$ is the  coordinate of the $n$-th viewpoint. $\mathcal{L}_{\mathrm{mot}}$ ensures that the motion posture of the 4D object's position at each moment aligns with the motion posture of the front view.

Merely optimizing the motion alignment of the 4D object with the video might cause the generated target's geometric structure to deviate from the 3D target. Therefore, it is essential to ensure that the 4D target maintains geometric consistency with the 3D input.
For unknown viewpoints of 4D object, another critical prior is the rendered images from the 3D representation at the same viewpoint. Although the rendered images of 3D and 4D from the same viewpoint may differ due to motion over time, their geometric structures remain highly correlated. By leveraging these same-view-rendered images from the 3D representation as geometric priors, we can ensure better alignment between the 4D output and the 3D target, avoiding undesirable modifications or artifacts introduced by the multiview diffusion model. This enhances the fidelity of the 4D geometry to the given 3D target. So, we propose $\mathcal{L}_{\mathrm{geo}}$ for optimization:
\begin{equation}
    \mathcal{L}_{\mathrm{geo}} = \frac{1}{N}\sum_{n=1}^{N}w(\tau)||\epsilon_{\phi}(\alpha_{\tau}x_{\theta_{t}}^{c_{n}}+\sigma_{\tau} \epsilon; x_{\psi}^{c_{n}} , 0,  d_{\text{MA}},\tau)-\epsilon||^{2}_{2}.
    \label{eq:geo}
\end{equation}
Since both \( x_{\theta_{t}}^{c_{n}} \) and \( x_{\psi}^{c_{n}} \) are rendered from the same viewpoint \( c_{n} \), the coordinate parameter in Equation~\ref{eq:geo} is zero. Thus, \( x_{\psi}^{c_{n}} \) is regarded  as the hypothetical front viewpoint and is utilized to refine the rendered image \( x_{\theta_{t}}^{c_{n}} \) of the 4D model through SDS optimization, following motion deformation, from the viewpoint $c_{n}$.
Here, \(x_{\psi}^{c}\) represents the rendered image of the 3D input $\psi$ from the same viewpoint \(c\). Additionally, considering that the structure of a 4D object can gradually diverge more from the 3D geometry over time, we introduce a temporal gradient coefficient $\frac{\mathcal{T}-t}{\mathcal{T}}\lambda$, where $\lambda$ is a hyperparameter used to control the scales of $\mathcal{L}_{\mathrm{mot}}$ and $\mathcal{L}_{\mathrm{geo}}$.
At this point, we can derive the $\mathcal{L}_{\mathrm{USVA}}$ to optimize unknow spatiotemporal viewpoint expression of 4D object:
\begin{equation}
    \mathcal{L}_{\mathrm{USVA}}=\frac{1}{\mathcal{T}}\sum^{\mathcal{T}}_{t=1}\frac{t}{\mathcal{T}}\lambda\mathcal{L}_{\mathrm{mot}}+\frac{\mathcal{T}-t}{\mathcal{T}}\lambda\mathcal{L}_{\mathrm{geo}}.
    \label{eq:USVA}
\end{equation}
From the above description, we can now derive the overall optimization objective $\mathcal{L}_{\mathrm{MGJA}}$:
\begin{equation}
    \mathcal{L}_{\mathrm{MGJA}} = \mathcal{L}_{\mathrm{KSVA}} +\mathcal{L}_{\mathrm{USVA}}.
    \label{eq:MGJA}
\end{equation}

\subsection{Asynchronous Optimization}\label{sec:AO}
The optimization objective $\mathcal{L}_{\mathrm{MGJA}}$ aims to refine the 4D representation by integrating both known and unknown temporal perspectives to align effectively with video and 3D data. However, employing a synchronous optimization approach, as seen in previous works~\cite{dreamgaussin4d}, can introduce instability, potentially leading to suboptimal generation results. To overcome this challenge, we propose an asynchronous optimization framework. As illustrated in Figure~\ref{fig:overall}, our method strategically alternates between fixing the 3DGS and the deformation network while optimizing the other. This asynchronous strategy ensures that when the 3DGS geometry is suboptimal, the deformation network can compensate. When the deformation network's capacity to drive motion is inadequate, the 3DGS reinforces the geometric structure. This approach fosters a more balanced integration of geometry and motion.

\begin{miao}

\begin{figure}[tp]
    \centering
    \includegraphics[width=1\linewidth]{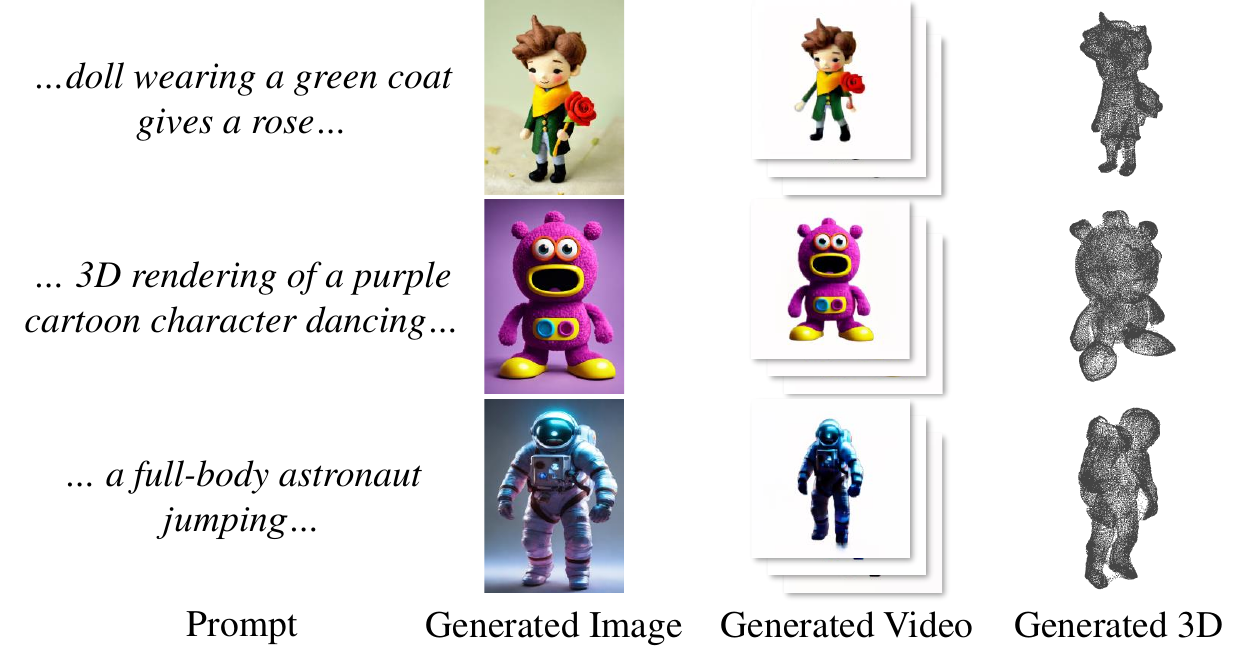}
    \caption{\textbf{Samples from X4D dataset.} Each quadruplet consists of a prompt, a generated image, a generated video, and a generated 3D object, all created by off-the-shelf diffusion models.}
    \label{fig:data}
    \vspace{-10pt}
\end{figure}

\subsection{X4D Dataset}\label{sec:data}
Figure~\ref{fig:data} presents representative samples from our X4D dataset, which is built from four types of aligned multimodal quadruplets (text, image, video, and 3D). We construct the dataset through four complementary pipelines:
(a) \textbf{Prompt-driven pipeline.} A textual prompt (from MAV3D~\cite{make-a-video-3d}) is first used to generate an image via SDXL~\cite{sdxl}. The same prompt then drives SVD~\cite{Stablevideo} to produce a video, and LGM~\cite{tang2024lgm} to reconstruct the corresponding 3D Gaussian representation.
(b) \textbf{Image-driven pipeline.} Given a single input image, we retrieve visually related content from the web, synthesize a short video using SVD, reconstruct its 3D Gaussian model with LGM, and produce a paired textual description using image-to-prompt tool~\cite{imagetoprompt}.
(c) \textbf{Video-driven pipeline.} For videos collected from Kling~\cite{Kling}, we extract the first frame as the reference image for 3D reconstruction. Each video is also processed by an image-to-text model to obtain its textual prompt.
(d) \textbf{3D-driven pipeline.} 3D assets from an open-source repository~\cite{meshy_ai_2025} are rendered from canonical viewpoints. The rendered image is then used for video synthesis and textual description generation, forming a complete text–image–video–3D quadruplet.
Together, these four pipelines produce consistently aligned multimodal quadruplets, enabling the creation of a large-scale, diverse dataset tailored for advancing 4D generation research.

The construction of X4D inevitably inherits statistical biases from its upstream generative components—SDXL~\cite{saharia2022photorealistic}, SVD~\cite{Stablevideo}, and LGM~\cite{tang2024lgm}—each of which is trained on heterogeneous data sources.
To mitigate cross-modal distribution mismatch, we intentionally combine models originating from diverse training corpora, with prompts generated via an image-to-prompt tool~\cite{imagetoprompt}. All assets, including text, images, videos, and 3D models, are collected from openly licensed community datasets.
During cross-modal expansion using generative tools such as SDXL and Kling, all generated samples undergo strict manual filtering to ensure data quality and regulatory compliance.

\end{miao}

\begin{figure*}[thbp]
    \centering
    \includegraphics[width=1\linewidth]{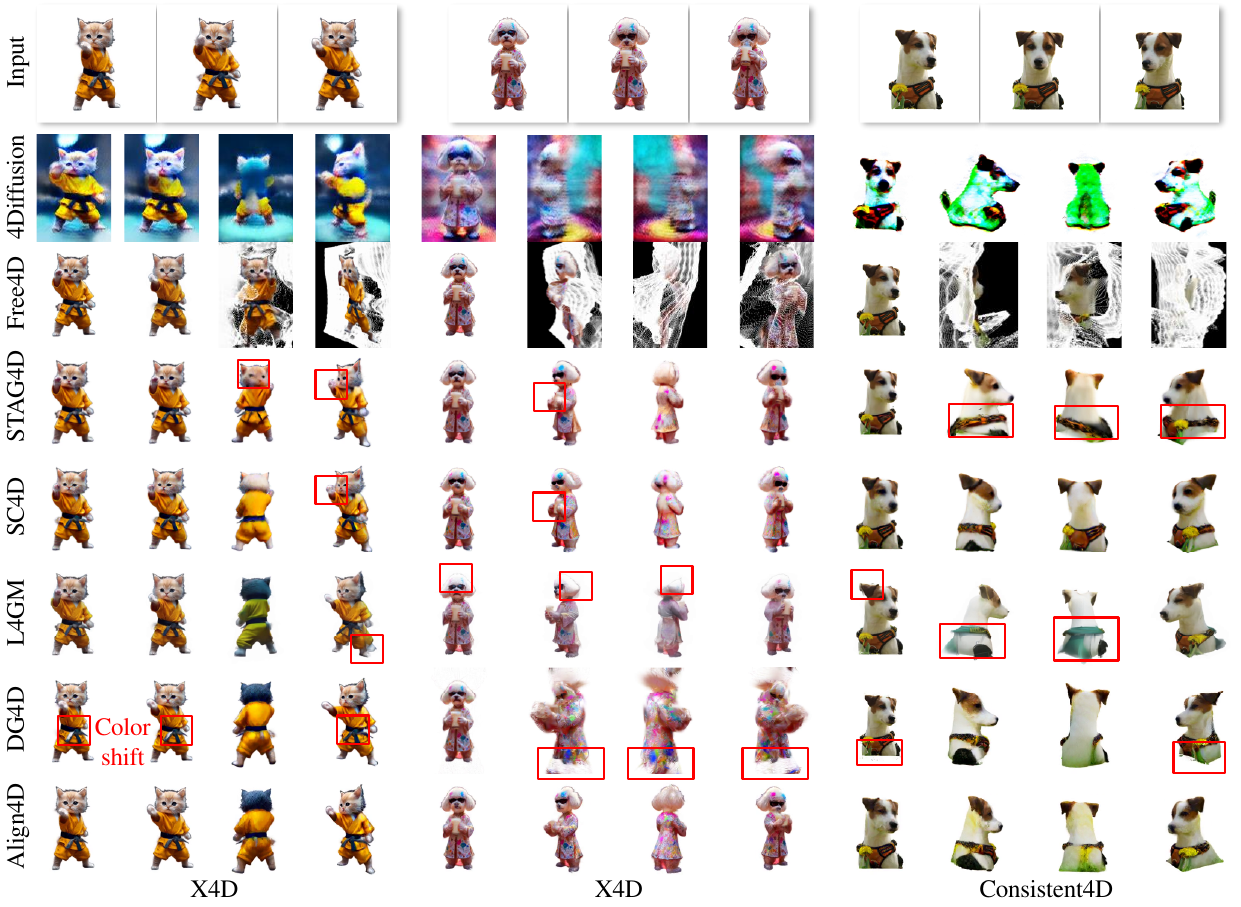}
    \vspace{-10pt}
    \caption{\textbf{Qualitative comparisons between Align4D and other methods on our X4D and Consistent4D datasets.} Align4D is capable of generating detailed, geometrically consistent 4D targets that faithfully replicate the motion observed in the video. To better appreciate these qualities, please zoom in. }
    \label{fig:data1}
    \vspace{-10pt}
\end{figure*}

\section{Experiment}
\subsection{Experimental Settings}

\textbf{Implementation Details.} 
In Align4D, we set the dense percentage to 0.1, the densification interval to 100, and the densification gradient threshold to 0.05. \miaot{All other settings follow DG4D~\cite{dreamgaussin4d}. In particular, DG4D uses a fixed object distance of 1.5 to align with the video, and sets the multi-view diffusion model parameter to 0 for SDS optimization, which we also adopt in our ablation experiments.}
During the object distance search, we set $d_{\text{min}} = 0.00001$ and $d_{\text{max}} = 3.00001$, performing searches at intervals of 0.05. The number of viewpoints is set to $N=4$, and we use Zero123~\cite{Zero-1-to-3} as the multiview diffusion model.
All experiments are conducted on a system equipped with NVIDIA V100 GPUs with 32 GB memory each, an Intel Xeon processor (Skylake, IBRS), and 629 GB of RAM. The software environment includes Ubuntu 22.04.3 LTS with CUDA 12.4.

\textbf{Dataset.} 
We conduct experiments on the proposed X4D dataset, which contains aligned text–image–video–3D quadruplets generated through our unified multimodal pipeline, as shown in Figure~\ref{fig:idea}. These quadruplets provide consistent input conditions across tasks, allowing fair comparisons among text-to-4D, image-to-4D, video-to-4D, and 3D-to-4D generation. To further evaluate performance, we also use the Consistent4D dataset~\cite{Consistent4d}, specifically designed for video-to-4D task. Both quantitative and qualitative experiments on X4D and Consistent4D assess the geometric consistency, motion coherence, and fidelity of the generated 4D outputs relative to the input control conditions.

\textbf{Metrics.}
We evaluate generation quality using PSNR~\cite{psnr}, SSIM~\cite{ssim}, LPIPS~\cite{lpips}, CLIP similarity~\cite{clip}, and FVD~\cite{fvd}, following established practice in 4D generation.
PSNR, SSIM, LPIPS, and CLIP similarity are adopted as image-level metrics to assess perceptual and semantic alignment between rendered images and references. FVD serves as a video-level metric widely used in generative video evaluation, capturing both frame-level fidelity and temporal coherence.
\begin{miao}
Additionally, we incorporate the CLIP-F score~\cite{bai2025syncammaster}, defined as the average CLIP similarity between adjacent frames, to evaluate 4D rendering and quantify temporal semantic consistency. We further incorporate VBench~\cite{vbench} to assess 360° surround-view renderings. VBench provides model-based scores along four dimensions—subject consistency, background consistency, aesthetic quality, and imaging quality—allowing a comprehensive evaluation of rendering quality.
\end{miao}

\textbf{Baselines.}
To ensure fair cross-modal comparison, all baselines are adapted to our unified X-to-4D framework, receiving their respective modality from the structured X4D quadruplets as input. This allows diverse methods to generate the same 4D targets under consistent conditions.
We include representative image-to-4D models~\cite{dreamgaussin4d,4dgen,free4D} and state-of-the-art video-to-4D methods~\cite{wu_sc4d_2024,Stag4d,ren2024l4gm,pan2024efficient4d,4Diffusion}. Text-to-4D~\cite{4d-fy} is excluded due to substantial differences in motion and geometric control. The standard 3D-to-4D method~\cite{Animate124} is impractical due to extreme memory demands; instead, we use 3D tools like CRM~\cite{Crm} and MeshyAI~\cite{meshy_ai_2025} with rigging for a feasible comparison.
\begin{miao}
Video generation models such as Kling~\cite{Kling} are additionally included, producing sequences under identical text or image conditions to ensure evaluation consistency.
\end{miao}
All methods are evaluated on X4D using the same quadruplets, ensuring fair comparison across modalities. We also report results on the Consistent4D dataset~\cite{Consistent4d}, which provides curated sequences for standardized assessment of motion fidelity, appearance consistency, and adherence to input conditions.

\begin{table*}[tp]
    \centering
    \caption{Quantitative results of Align4D on our X4D dataset.}
\resizebox{1\linewidth}{!}{
    \begin{tabular}{c|cccc|cccc}
    \toprule
          \multirow{2}{*}{Methods} & \multicolumn{4}{c}{Human Evaluation}  & \multicolumn{4}{c}{VBench} \\ 
          & Appearance\% $\uparrow$& Structure\% $\uparrow$& Motion\% $\uparrow$ & Fidelity\% $\uparrow$ & \makecell{Subject \\ Consistency $\uparrow$}   & \makecell{Bckground \\Consistency $\uparrow$}  & \makecell{Aesthetic \\ Quality $\uparrow$}   & \makecell{Imaging \\ Quality $\uparrow$} \\ 
    \midrule
    L4GM~\cite{ren2024l4gm} & 9.2 & 5.5  & 7.9  & 6.6  &0.80&0.88&0.46&0.40\\
    SC4D~\cite{wu_sc4d_2024} & 5.3 & 5.3  & 4.0  & 5.3  &0.81&0.91&0.43&0.38\\
    STAG4D~\cite{Stag4d} & 19.7 & 17.1  & 26.3  & 18.4  &0.83&0.91&0.50&0.41\\
    DG4D~\cite{dreamgaussin4d} (baseline) & 4.0  & 2.6  & 3.9  & 3.9  &0.71& 0.85&0.40&0.31\\
    \midrule
    Align4D (ours)  & \textbf{61.8} & \textbf{69.5} & \textbf{57.9} & \textbf{65.8}  & \textbf{0.85}& \textbf{0.93}& \textbf{0.55} & \textbf{0.43}\\
    \bottomrule
    \end{tabular}}
    \label{tab:x4d}
\end{table*}

\begin{table}[tp]
    \centering
    \caption{Quantitative results of Align4D on the Consistent4D dataset. The best scores are highlighted in \textbf{bold}.
$*$ indicates generating videos using the first and last frames of the training video.
$+$ indicates generating videos using the first and last frames of the test video.}

\resizebox{1\linewidth}{!}{
    \begin{tabular}{c|cccccc}
    
    \toprule
          \multirow{2}{*}{Methods} & \multicolumn{5}{c}{Consistent4D Dataset~\cite{Consistent4d}}   \\ 
          & PSNR$\uparrow$ & SSIM$\uparrow$ & LPIPS$\downarrow$ & FVD$\downarrow$ & CLIP$\uparrow$ & CLIP-F$\uparrow$\\ 
    \midrule
    CRM~\cite{wang2024crm} + Rigging &13.5	&0.88	&0.17	&1258.5	&0.79 &-\\
    Meshyai~\cite{meshy_ai_2025} + Rigging &12.7	&0.87	&0.19	&1179.2	&0.85 &-\\
    Kling$^{*}$~\cite{Kling} &11.0	&0.81	&0.28		& - &0.81 &-\\
    Kling$^{+}$~\cite{Kling} &11.4	&0.80	&0.26		& - &0.78 &-\\
    \midrule
    4Diffusion~\cite{4Diffusion} & 12.2 & 0.84 & 0.26 & 1522.9 & 0.83 & 0.912\\ 
    Free4D~\cite{free4D} & 6.4 & 0.45 & 0.41 & 2513.6 & 0.77 & 0.809\\ 
    L4GM~\cite{ren2024l4gm} & 14.2 & 0.84 & 0.20 & 1217.1 & 0.90 & 0.990\\ 
    SC4D~\cite{wu_sc4d_2024} & 16.8 & 0.86 & 0.16 & 1132.2 & 0.91 & 0.988\\ 
    4DGen~\cite{yin20234dgen} &12.7	&0.87	&0.19	&1258.5	&0.80 & 0.981\\
    Efficient4D~\cite{pan2024efficient4d} &12.8	&0.85	&0.21	&1304.2	&0.89 &0.954\\
    STAG4D~\cite{Stag4d} & 17.0 & 0.87 & 0.14 & 1251.7 & 0.90 &0.988\\ 
    DG4D~\cite{dreamgaussin4d} (baseline) & 10.7 & 0.78& 0.28 & 1262.0 & 0.89 & 0.978\\ 
    \midrule
    Align4D (ours)  & \textbf{17.8} & \textbf{0.90} & \textbf{0.11} & \textbf{1088.9} & \textbf{0.94} & \textbf{0.992}\\ 
    
    \bottomrule
    \end{tabular}}
    
    \label{tab:c4d}
\end{table}

\begin{table}[tp]
    \caption{Quantitative results from ablation studies of Align4D on Consistent4D dataset. }
    \centering
\resizebox{1\linewidth}{!}{
    \begin{tabular}{l|ccccc}
    \toprule
          \multirow{2}{*}{Methods} & \multicolumn{5}{c}{Consistent4D Dataset~\cite{Consistent4d}}   \\ 
          & PSNR$\uparrow$ & SSIM$\uparrow$ & LPIPS$\downarrow$ & FVD$\downarrow$ & CLIP$\uparrow$ \\ 
    \midrule
    STAG4D~\cite{Stag4d} & 	17.0	&0.87	&0.14	&1251.7	&0.90 \\ 
    STAG4D~\cite{Stag4d} + ODA + MGJA & 17.5	&0.88	&0.12	&1109.4	&0.93 \\ 
    STAG4D~\cite{Stag4d} + AO & 17.2	&0.88	&0.14	&1129.2	&0.91 \\ 
    
    \midrule
    DG4D~\cite{dreamgaussin4d} (baseline) & 10.7 & 0.78& 0.28 & 1262.0 & 0.89 \\ 
    Align4D W/O. ODA  & 15.9 & 0.86& 0.16 & 1111.5  & 0.90 \\ 
    Align4D W/O. MGJA & 15.5    & 0.85& 0.17 & 1113.1 & 0.91 \\ 
    Align4D W/O. AO   & 17.6 & 0.89 & 0.13 & 1139.7 & 0.93 \\ 
    Align4D (ours)  & \textbf{17.8} & \textbf{0.90} & \textbf{0.11} & \textbf{1088.9} & \textbf{0.94} \\ 
    
    \bottomrule

    \end{tabular}}
    \label{tab:bal}
    \vspace{-10pt}
\end{table}

\subsection{Comparisons to State-of-the-Art Methods}
\textbf{Qualitative evaluation.}  
We provide extensive visual comparisons on the X4D dataset, where the input data is generated by pretrained generative models. As shown in Figure~\ref{fig:data1}, Align4D produces high-quality, sharp renderings with strong temporal and multiview consistency.  
\miaot{In comparison, 4Diffusion~\cite{4Diffusion}, despite being a large-scale multiview video diffusion model, struggles to generalize to the diverse test scenarios. Free4D~\cite{free4D}, while integrating the powerful 4D reconstruction network MonST3R~\cite{MonST3R}, often fails to generate geometrically detailed and temporally coherent 4D targets that are viewable from arbitrary angles.  
STAG4D~\cite{Stag4d} generally produces geometric artifacts, whereas SC4D~\cite{wu_sc4d_2024} reduces geometric errors but lacks fine-grained detail. L4GM~\cite{ren2024l4gm} achieves fast generation, yet exhibits noticeable geometric inconsistencies. DG4D~\cite{dreamgaussin4d} can approximate the frontal-view video visually; however, it occasionally deviates from the video or introduces substantial errors in other viewpoints.  }
By contrast, Align4D consistently delivers the most refined 4D outputs, preserving dynamic motion and geometric fidelity across all views.

\begin{miao}
\textbf{Quantitative evaluation.} 
We first conduct comparisons on the X4D dataset. All models generate 4D targets conditioned on the same quadruplet inputs and render full 360$^\circ$ surround-view videos. A user study with 30 participants evaluates the generated results across four dimensions: Appearance, Structure, Motion, and Fidelity. Align4D consistently receives the highest human preference. Additionally, the rendered videos are evaluated using VBench on subject consistency, background consistency, aesthetic quality, and imaging quality, where Align4D also achieves the top scores.
Furthermore, Table~\ref{tab:c4d} reports the quantitative results on the Consistent4D dataset. Compared to state-of-the-art video-to-4D methods~\cite{ren2024l4gm, wu_sc4d_2024, Stag4d, 4dgen, pan2024efficient4d, dreamgaussin4d}, Align4D attains the best performance across five key metrics, covering both image-level and video-level evaluations. This demonstrates that Align4D effectively integrates the geometric consistency of 3D data with the motion coherence of video data, successfully transferring these properties to novel, unseen views. 
In contrast, 3D generation models~\cite{Crm, meshy_ai_2025} combined with manual rigging struggle to faithfully reproduce object motion, and video generation models~\cite{Kling} fail to maintain motion consistency under test-view conditions. Overall, Align4D exhibits clear advantages over these alternatives, producing more accurate, temporally coherent, and geometrically consistent 4D results.
\end{miao}

\begin{figure*}[htp]
    \centering
    \includegraphics[width=1\linewidth]{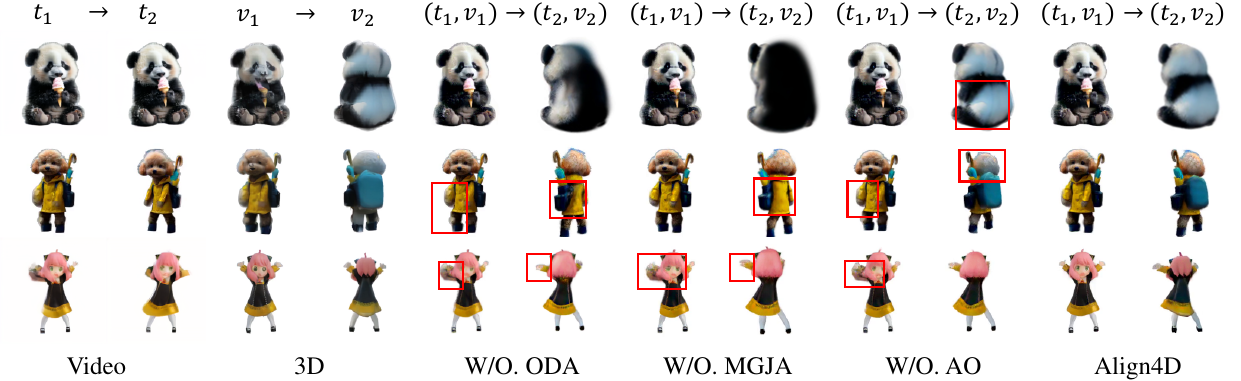}
    \caption{\textbf{Qualitative ablation experiments on the X4D dataset for Align4D.} ODA is crucial for precisely injecting action and geometric data into video and multiview diffusion models. MGJA significantly enhances the rendering of 4D non-frontal, multi-moment views by meticulously aligning 4D actions to video and geometry to 3D inputs. Additionally, AO refines the intricate details of the generated 4D targets.}
    \label{fig:abl}
    \vspace{-8pt}
\end{figure*}

\begin{table}[tp]
    \caption{ Comparison of loss reduction between Asynchronous Optimization (AO) and Joint Optimization (JO).}
    \centering
\resizebox{0.9\linewidth}{!}{
    \begin{tabular}{c|ccccc}
    \toprule
    \diagbox{Methods}{Step} & 100 & 200 &300 & 400 & 500 \\ 
    \midrule
    Align4D + JO & 72.3 &174.2 &155.2 &44.0 &3.5 \\
    Align4D + AO &34.4 &87.7 &85.7 &24.8 &3.2 \\
    \bottomrule
    \end{tabular}}
    \label{tab:AOJO}
\vspace{-8pt}
\end{table}

\begin{table}[tp]
    \centering
    \caption{Runtime comparison between Align4D and baseline methods. Only the time spent on generating 4D targets is measured, while the overhead for data preparation is excluded.}
\resizebox{0.75\linewidth}{!}{
    \begin{tabular}{c|c|c}
    
    \toprule
    Methods & Time (Min.) & VRAM (GB) \\
    \midrule
     L4GM~\cite{ren2024l4gm}  & 1.5  & 24.6 \\
     SC4D~\cite{wu_sc4d_2024}& 35  & 8.9\\
     STAG4D~\cite{Stag4d} & 80 & 12.2\\
     DG4D~\cite{dreamgaussin4d} & 15 & 15.6\\ 
     \midrule
     Align4D (ours) & 25  & 19.4\\  
     ODA-VAOD & 0.0025 & 2.7\\ 
     ODA-MAOD & 0.20 & 4.5\\  

    \bottomrule
    \end{tabular}}
    \label{tab:time2}
    \vspace{-6pt}
\end{table}

\subsection{Ablation Studies} 
\textbf{The qualitative results are shown in Figure~\ref{fig:abl}}. As can be seen, the absence of object distance alignment leads not only to a blurred frontal view but also causes the back view to deviate from the 3D data priors when not using ODA for SDS optimization, resulting in erroneous coverage modifications. Without MGJA, unknown views, especially the rear view of the 4D generation target, exhibit more severe errors and inconsistencies. If asynchronous optimization is not used, although there is good alignment of dynamics and geometry with the input video and 3D data in both frontal and other views, some details lack refinement and there is a tendency to inherit imperfections from imperfect video or 3D data. Asynchronous optimization effectively alleviates these issues. When ODA, MGJA, and AO are used together, forming the complete Align4D, the 4D target achieves fidelity to video motion, 3D geometry, and exhibits detailed precision.

\textbf{The quantitative results are shown in Table~\ref{tab:bal}}. 
By synergistically employing ODA, MGJA, and AO, Align4D effectively addresses limitations in generated video-3D data, producing 4D targets with motion and geometry better aligned with human perception.
ODA is a prerequisite for robust MGJA, disabling it significantly degrades performance (Table~\ref{tab:bal} W/O. ODA). MGJA then refines motion and geometry alignment using ODA's precise object distances; the effectiveness hinges on ODA's quality (Table~\ref{tab:bal} W/O. MGJA). AO further enhances MGJA's alignment by decoupling 3DGS and deformable network optimization, yielding finer and more stable convergence by mitigating component interference (Table~\ref{tab:bal} W/O. AO).  \miaot{
Besides, Table~\ref{tab:time2} reports the computational costs of Align4D and baseline methods, highlighting the low overhead of VAOD and MAOD searches. 
In summary, the ODA module estimates accurate object distances, MGJA ensures robust multiview and temporal alignment, and AO facilitates optimal convergence. Together, these components allow Align4D to achieve state-of-the-art generation quality with balanced computational efficiency.
}

\textbf{Module migratability.} We successfully migrate our modules to the STAG4D method. We specifically integrate ODA, MGJA, and AO into STAG4D (see Table~\ref{tab:bal}). Notably, MGJA's synergistic function relies on the optimal object distance provided by ODA; thus, these two modules are inherently bound together for migration. This demonstrates that ODA, MGJA, and AO from Align4D can be directly transferred to other models like STAG4D, providing performance gains across multiple metrics compared to the STAG4D baseline.

\textbf{Asynchronous optimization effectiveness.}
We further compared the effects of joint optimization (JO) and asynchronous optimization (AO) during the training phase to highlight their differences. As shown in Table~\ref{tab:AOJO}, under the same number of steps, asynchronous optimization resulted in smaller fluctuations and a faster decrease in loss, using an SDS loss as an example. It also achieved a lower loss value at the same number of steps. This indicates that decoupling the optimization of the deformation network and 3DGS is more beneficial for aligning 4D objects with the input conditions.

\begin{figure}[tp]
    \centering
    \includegraphics[width=0.85\linewidth]{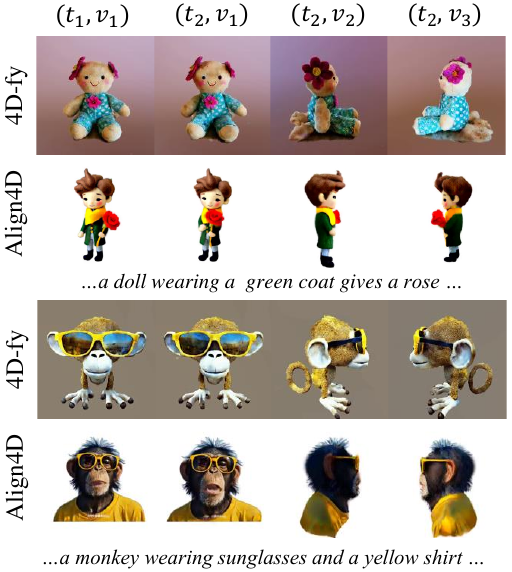}
    \caption{\textbf{Generation results comparing Align4D with text-to-4D method~\cite{4d-fy}.} Given the same text prompts, Align4D delivers markedly superior geometric structure and a significantly wider motion range.}
    \label{fig:4dfy}
\end{figure}

\subsection{Comparison with Text-to-4D generation method}

Compared to the text-to-4D generation method 4D-Fy~\cite{4d-fy}, the dynamic quality of 4D outputs controlled solely by text prompts is noticeably inferior to those guided by images or videos, even when using the same prompts, as illustrated in Figure~\ref{fig:4dfy}. Moreover, text-to-4D methods cannot accept inputs from other modalities, such as images or videos, making both qualitative and quantitative evaluation on the Consistent4D dataset infeasible. 
In contrast, Align4D’s flexible X-to-4D generation pipeline not only supports a wide range of input modalities but also leverages the smooth motion from the generated video and the fine-grained geometric guidance provided by the generated 3D object.

\begin{figure}[tp]
    \centering
    \includegraphics[width=1\linewidth]{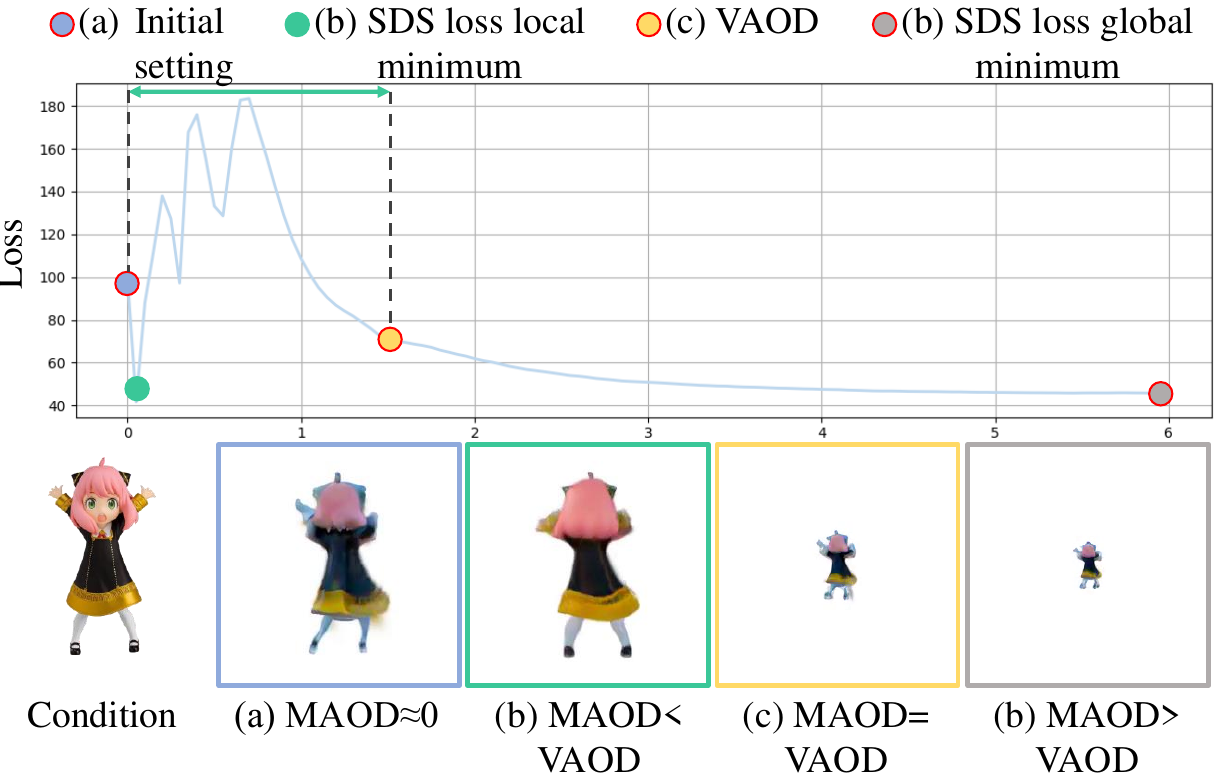}
    \caption{
        \miaot{
        \textbf{Generation results under different MAOD selections.}
(a) A small object distance (OD) used as MAOD introduces color shifts and loss of fine details.
(b) An SDS-loss local-minimum OD used as MAOD achieves the best input fidelity.
(c) Using VAOD as MAOD produces an under-scaled object with reduced high-frequency detail.
(d) An SDS-loss global-minimum OD used as MAOD likewise leads to under-scaling.
        }
        }
    \label{fig:ma}
\end{figure}

\begin{figure}[tp]
    \centering
    \includegraphics[width=1\linewidth]{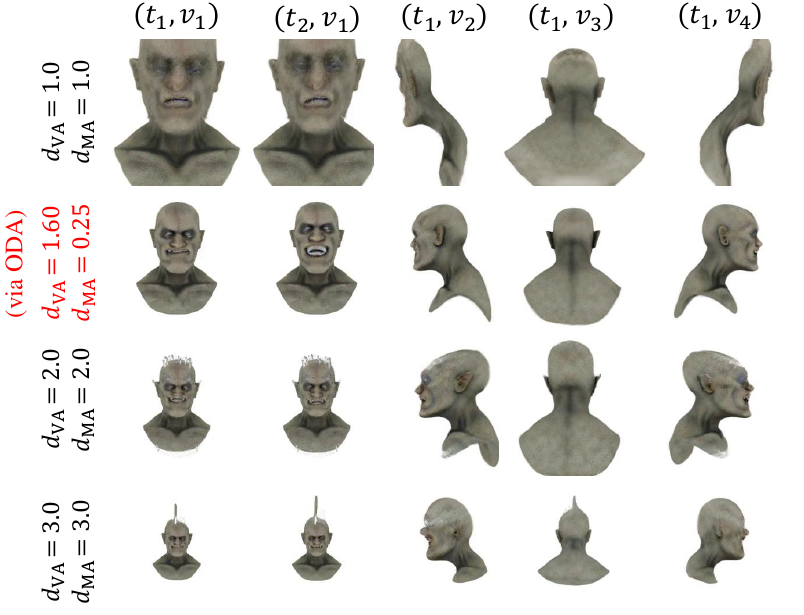}
    \caption{\textbf{4D generation results under different object distances.} Only when the optimal values are found using the object distance alignment method for $d_{\text{VA}}$ and $d_{\text{MA}}$, can we ensure the generation of 4D targets with faithful motion and accurate geometry.}
    \label{fig:FLS}
    \vspace{-10pt}
\end{figure}

\section{Disscussion}

\begin{miao}



\subsection{Why MAOD Should Be Smaller Than VAOD}\label{sec:why}

MAOD is designed to approximate the intrinsic object distance preferred by the multiview diffusion model, which often differs from the viewpoint-aligned object distance (VAOD). Figure~\ref{fig:ma} illustrates this effect by varying MAOD while keeping VAOD fixed:
(a) \textbf{MAOD $\approx 0$ (initial setting)}: The generated results exhibit local blurring and overall color shifts, indicating that the multiview diffusion model cannot accurately infer the target geometry under the current object distance setting.
(b) \textbf{MAOD $<$ VAOD (local SDS-loss minimum)}: The generated geometry is complete, and the rendered appearance closely matches the input, suggesting that this distance best aligns with the diffusion model’s implicit distance prior.
(c) \textbf{MAOD = VAOD}: The generated object appears underscaled with color inconsistencies, indicating that VAOD does not capture the diffusion model’s internal understanding of object distance, and thus MAOD is required for proper alignment.
(d) \textbf{MAOD $>$ VAOD (global SDS-loss minimum)}: Although this setting achieves the lowest SDS loss, it produces unrealistic object scale and appearance, demonstrating that global SDS-loss minimization does not correspond to a physically meaningful distance alignment.
These observations indicate that the diffusion model favors an object distance slightly smaller than VAOD, typically near a local SDS-loss minimum, making it the most appropriate choice for defining MAOD.

\end{miao}

\subsection{Generation Quality Comparison Across Object Distances} 
Figure~\ref{fig:FLS} shows a visual comparison of 4D generation results under different object distances. Using an unmatched video object distance causes noticeable blur in frontal-view actions and introduces conflicts with the multiview diffusion model supervision, resulting in geometric inaccuracies. Similarly, an unmatched multiview object distance leads to misalignment of the target geometry across viewpoints. By applying Object Distance Alignment (ODA), the matched object distances ensure consistent video motion and 3D object geometry, producing coherent and high-quality 4D assets.

\begin{miao}
\begin{figure*}[tp]
    \centering
    \includegraphics[width=1\linewidth]{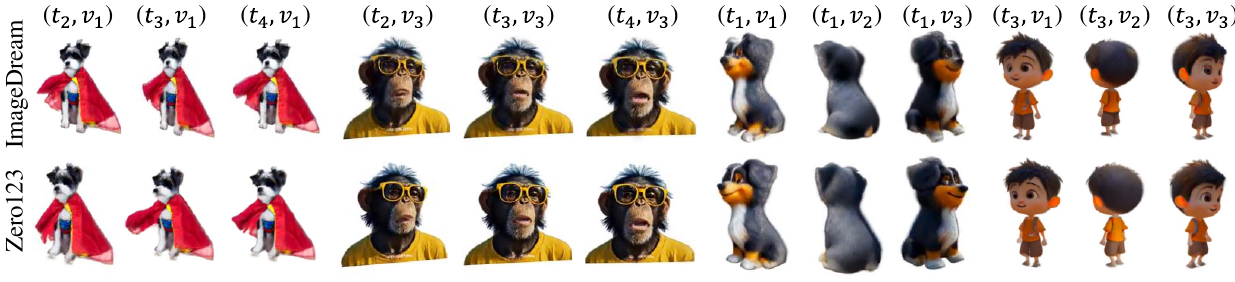}
    \caption{\textbf{Align4D with different multiview diffusion models.} 
We incorporate two distinct multiview diffusion models, Zero123~\cite{Zero123++} and ImageDream~\cite{imagedream}, both of which produce outputs with accurate geometric structures and temporally coherent motion. These results further demonstrate that Align4D is compatible with a variety of multiview diffusion backbones.}
    \label{fig:diff-diff}
\end{figure*}

\begin{figure}[tp]
    \centering
    \includegraphics[width=0.7\linewidth]{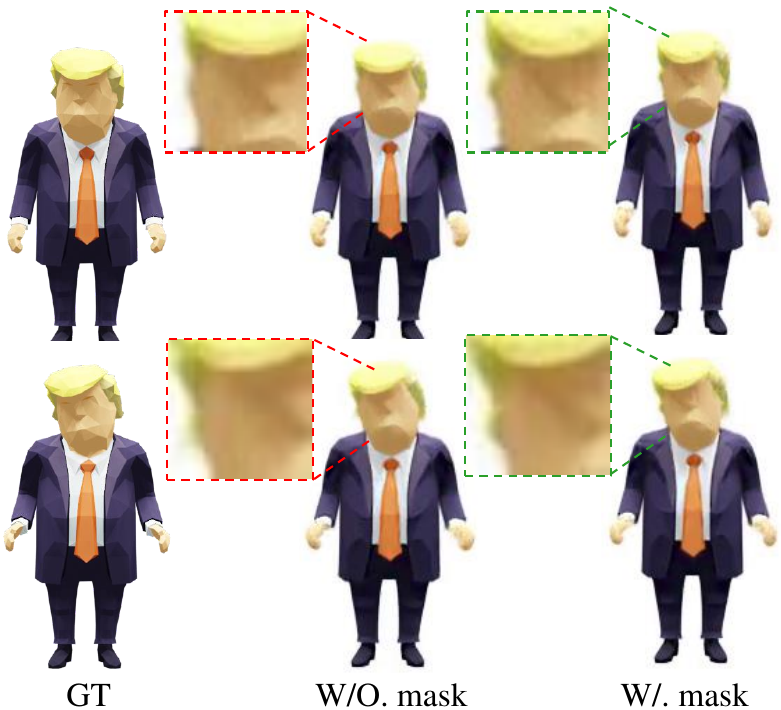}
    \caption{\textbf{Effect of masks on 4D generation results.} Incorporating masks further enhances boundary fidelity and preserves fine edge details in the generated 4D content.}
    \label{fig:mask2}
\end{figure}

\subsection{Effect of Masks}
Following DG4D~\cite{dreamgaussin4d}, we employ RemoveBg~\cite{removebg}, based on a pretrained U$^{2}$Net~\cite{u2net}, to extract frame-wise masks that guide generation toward better alignment with the input video geometry. Figure~\ref{fig:mask2} presents an ablation study on the mask loss. The results indicate that Align4D can generate high-quality geometry and maintain motion consistency even without masks. However, incorporating masks improves the preservation of fine local details and achieves a closer visual match to the input, particularly in the frontal view.

\begin{figure}[tp]
    \centering
    \includegraphics[width=0.95\linewidth]{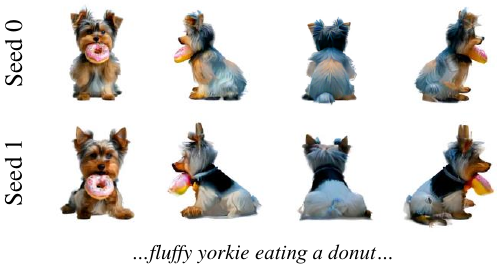}
    \caption{\textbf{Visualization of diversity in generated results shows that by using different random seeds.} Align4D can produce 4D outputs with significant variations while faithfully adhering to the input conditions.}
    \label{fig:diversity}
\end{figure}

\subsection{Stability of Generation Across Different Diffusion Models}
MAOD exhibits broad compatibility with various diffusion models. Methods such as Zero123~\cite{Zero-1-to-3} allow explicit control of the object distance by directly using camera parameters as inputs. In contrast, multiview diffusion models represented by ImageDream~\cite{imagedream} employ a fixed distance schedule: they rely on several predefined camera parameters to generate multiview outputs, and their pretrained weights implicitly encode a fixed viewing distance. Consequently, only an extremely lightweight modification to Equation~\ref{eq:core} is required for adaptation to these models.
Specifically, within the distance range $\mathcal{D}=[d_{\text{min}}, d_{\text{max}}]$, we render image sets 
$\left\{ \left\{ x_{\theta_{1}}^{c,d'} \right\}_{c \in C} \right\}_{d' \in \mathcal{D}}$,
where $c$ corresponds to three horizontally varying azimuth angles $[-90^\circ, 0^\circ, 90^\circ]$,
which lie within the azimuth range covered by the ImageDream training data.
The SDS loss for each candidate distance $d'$ is computed as:
\begin{equation}
\mathcal{L}_{\text{SDS}}^{d'} = \frac{1}{|C|\,|\text{T}|}\sum_{c \in C}\sum_{\tau \in \text{T}}w(\tau)\left\|\epsilon_{\phi}\!\left(z_{\theta_{1}}^{c,d'} ; I_{1}, c, \tau\right)- \epsilon\right\|^{2}_{2}.
\end{equation}
We select the distance $d'$ that is smaller than VAOD and corresponds to a local minimum of the SDS loss as the MAOD. Figure~\ref{fig:diff-diff} presents results generated using different diffusion models. Align4D produces highly consistent outputs across multiple diffusion backbones and remains stable across diverse examples, demonstrating that MGJA in Align4D is broadly applicable to various multiview diffusion models.

\subsection{Align4D with Different Seeds}
To investigate the impact of random seeds on 4D generation, we use the text-to-4D task as an example, since it involves the most extensive data processing through multiple diffusion models. Given a fixed prompt, an image is first generated via SDXL, followed by video synthesis with SVD and 3D generation with LGM to construct a video–3D pair. Align4D then generates the corresponding 4D object. As shown in Figure~\ref{fig:diversity}, while different seeds introduce stylistic variations in the video–3D pair, the resulting 4D objects consistently preserve the input text semantics and maintain stable geometric structure across seeds.

\subsection{Robustness Across Diverse Video and 3D Generation Models}

Align4D operates on video–3D pairs generated by various diffusion models, regardless of the underlying architecture. Table~\ref{tab:rob} presents results obtained using different combinations of video and 3D generative models. For each configuration, the same conditioning input is used to synthesize the video–3D pair, which Align4D then processes to produce the final 4D output.  
A user study with 30 participants further evaluates these results. Across all combinations, Align4D receives comparable user preferences, demonstrating that the framework consistently delivers high-quality 4D generation, thereby further validating the model versatility of Align4D.

\begin{table}[tp]
    \caption{ User Preference for Results Generated by Different Models Integrated into the Align4D Pipeline }
    \centering
\resizebox{0.8\linewidth}{!}{
    \begin{tabular}{c|cc}
    \toprule
    \diagbox{3D model}{Video model} & 	Kling~\cite{Kling}	&VideoCrafter~\cite{videocrafter2} \\ 
    \midrule
    Meshai~\cite{meshy_ai_2025}  & 27.4\% & 22.5\% \\
    Tripo3d~\cite{Tripo3D}  & 26.5\% & 23.6\% \\
    \bottomrule
    \end{tabular}}
    \label{tab:rob}
\end{table}

\subsection{Failure Analysis}
\begin{figure}[tp]
    \centering
    \includegraphics[width=0.9\linewidth]{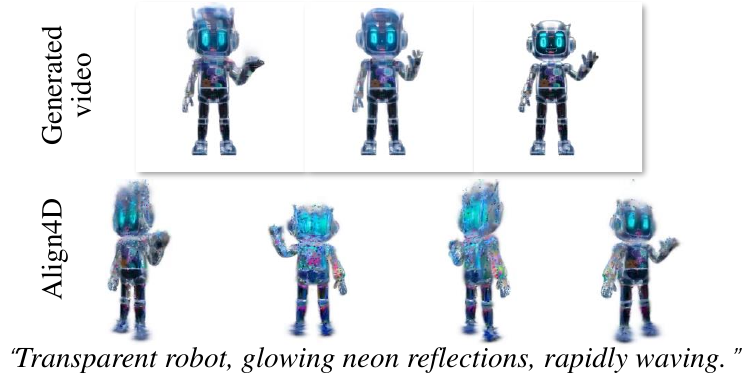}
    \caption{\textbf{Failure case.} The generated 4D assets exhibit limited fine detail when handling transparent or rapidly flickering neon objects.}
    \label{fig:fail}
\end{figure}

We further analyze scenarios where Align4D underperforms. When using a text-to-image model followed by an image-to-video model, with the resulting image-to-3D output to make a video-3D pair as input, the generated 4D assets inherit limitations from the underlying diffusion models. In particular, current text-to-image and image-to-video diffusion models struggle with transparent materials and rapidly changing neon effects. Consequently, Align4D’s 4D outputs also reflect these deficiencies. We expect that as generative models improve in expressiveness, these limitations will be mitigated.

\end{miao}

\section{Conclusion}
In this paper, we propose Align4D, an alignment-based framework designed for flexible X-to-4D generation. This framework leverages multimodal inputs combined with pre-trained diffusion models to produce paired video-3D  data . Align4D introduces object distance-based alignment, searching for the video-aligned object distance and the multiview-aligned object distance that achieves optimal generalization for multiview diffusion models. Based on this, the framework aligns the front-view renderings of 4D targets with the video and conditions unknown non-front view multi-timestep renderings on the same-timestep front-view video frames and the same-view initial-timestep 3D renderings. By employing SDS optimization, Align4D achieves joint alignment of motion and geometry. Furthermore, asynchronous optimization is utilized to refine the 4D target for better motion and geometry representation. To evaluate the open X-to-4D generation task, we propose a quadruplet dataset, X4D, consisting of (prompt, image, video, 3D). Through extensive testing in generated scenes, real-world scenarios, and synthetic environments, Align4D demonstrates exceptional generative capabilities.


\bibliography{IEEEabrv, ref}
\bibliographystyle{IEEEtran}
%

\begin{IEEEbiography}[{\includegraphics[width=1.0in, clip, keepaspectratio]{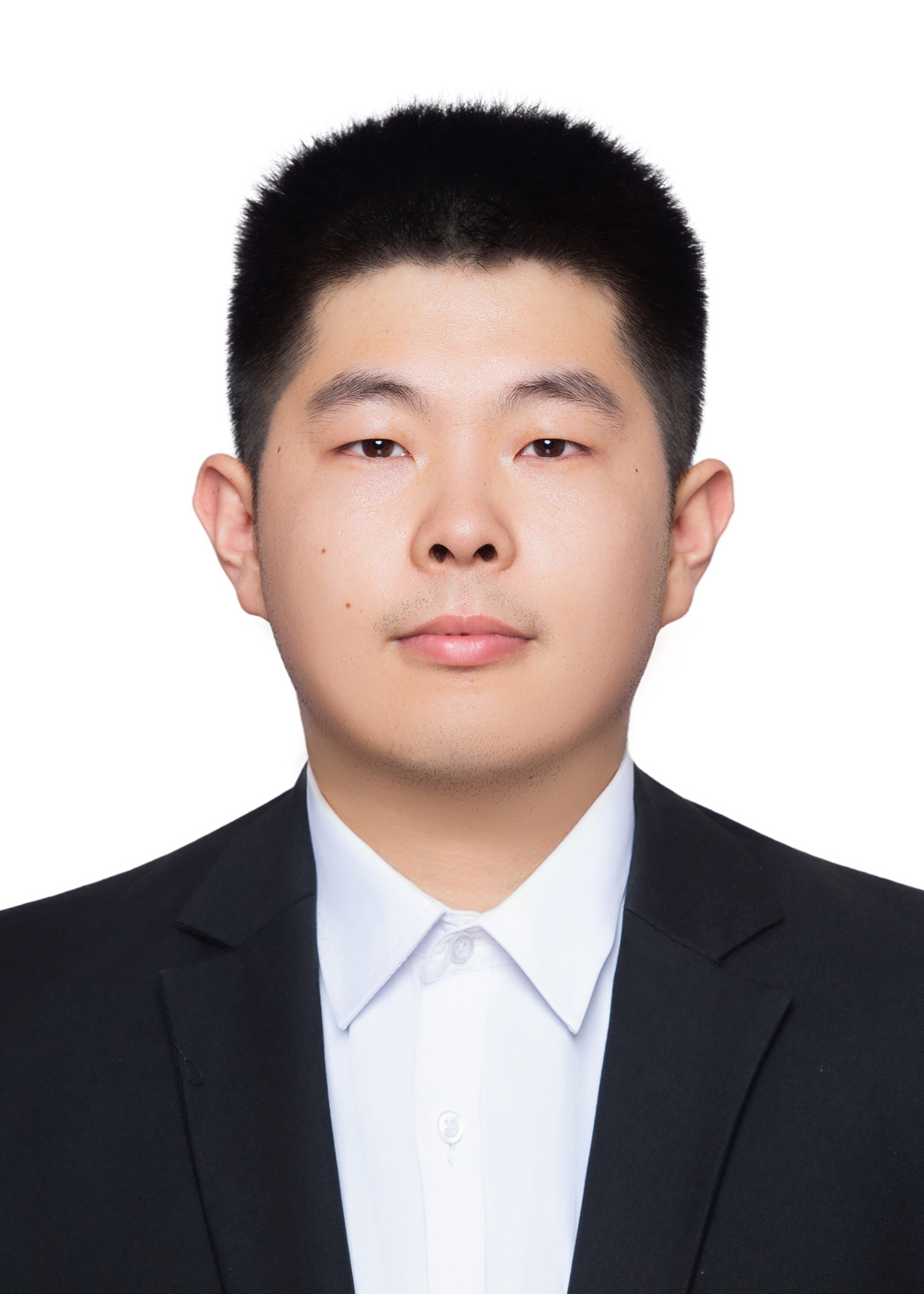}}]{Qiaowei Miao} received a bachelor’s degree in computer science and technology from Hebei University, China, in 2021. He is working
toward a PhD at the School of Software Technology, Zhejiang University, China. His research interests include 4D vision and deep learning.
\end{IEEEbiography}

\begin{IEEEbiography}[{\includegraphics[width=1.0in, clip, keepaspectratio]{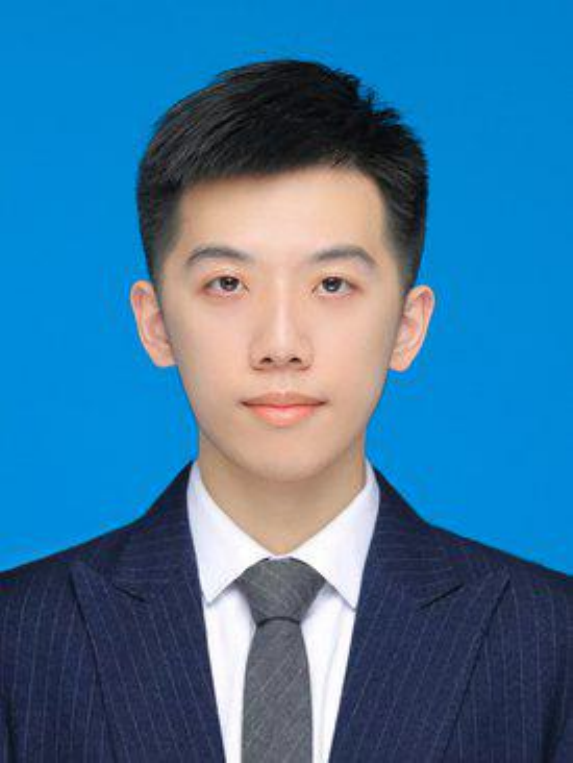}}]{Kehan Li}
received a bachelor's degree in computer science and technology at Chongqing University, China, in 2024. He is pursuing a master's degree at Zhejiang University, focusing on Artificial Intelligence and Computer Vision.
\end{IEEEbiography}

\begin{IEEEbiography}[{\includegraphics[width=1.0in, clip, keepaspectratio]{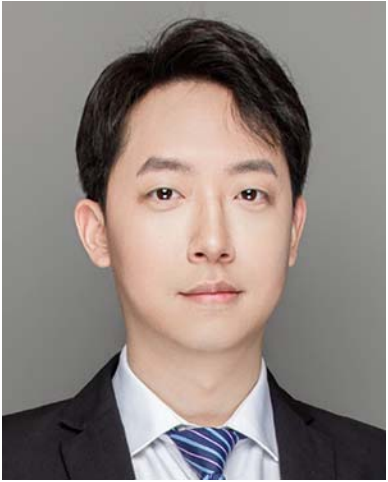}}]{Yawei  Luo}
received the PhD degree from the
Huazhong University of Science and Technology, in
2020. He is a ZJU 100 young professor with the
School of Software Technology, Zhejiang University.
He was a postdoctoral researcher with CCAI, College
of Computer Science and Technology in Zhejiang
University from 2020 to 2023. He was a visiting Ph.D
student with ReLER lab, AAII, University of Technology Sydney, from 2017 to 2019. His research interests include knowledge engineering, domain adaptation, and 3D reconstruction.
\end{IEEEbiography}

\begin{IEEEbiography}[{\includegraphics[width=1.0in, clip, keepaspectratio]{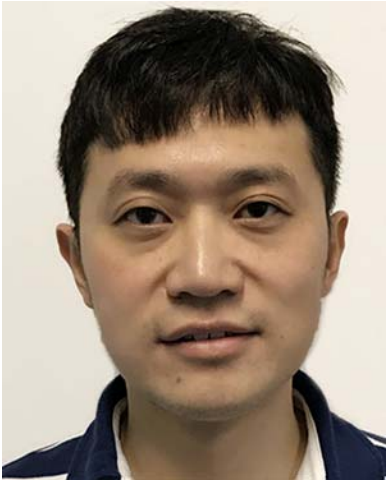}}]{Yi Yang}
received the PhD degree in computer science from Zhejiang University, Hangzhou, China,
in 2010. He is currently a distinguished professor
with Zhejiang University, China. He was a professor and director with the ReLER Lab, Australian
Artificial Intelligence Institute (AAII), University of
Technology Sydney, Australia. He was a postdoctoral research with the School of Computer Science,
Carnegie Mellon University, Pittsburgh, PA, USA.
His current research interest include machine learning
and its applications to multimedia content analysis
and computer vision, such as multimedia indexing and retrieval, surveillance
video analysis and video semantics understanding.
\end{IEEEbiography}

\vfill


\vfill

\end{document}